\documentclass[runningheads]{llncs}
\usepackage[T1]{fontenc}
\usepackage[colorlinks, citecolor=blue]{hyperref}
\usepackage{graphicx}
\graphicspath{ {img/} }
\usepackage{amsmath}
\usepackage{amssymb}
\usepackage{anyfontsize}
\usepackage{booktabs,etoolbox}
\usepackage{xcolor}
\usepackage{subfig}
\usepackage{multirow}
\usepackage{mdframed}
\newmdenv[leftline=false,rightline=false,linewidth=.8]{topbot}

\newcommand{\myparagraph}[1]{\paragraph{\emph{\textbf{#1}}}}
\definecolor{tum}{RGB}{48, 112, 179}
\DeclareMathOperator*{\argmax}{arg\,max}
\DeclareMathOperator*{\argmin}{arg\,min}
\begin{document}
\title{3D Arterial Segmentation via Single 2D Projections and Depth Supervision in Contrast-Enhanced CT Images }
\titlerunning{3D Arterial Segmentation via Single 2D Projections and Depth Supervision}

\author{Alina F. Dima\inst{1,2} \and
Veronika A. Zimmer\inst{1,2} \and
Martin J. Menten\inst{1,4} \and
\mbox{Hongwei Bran Li\inst{1,3}} \and
Markus Graf\inst{2} \and
Tristan Lemke\inst{2} \and
Philipp Raffler\inst{2} \and
\mbox{Robert Graf\inst{1,2}} \and
Jan S. Kirschke\inst{2} \and
Rickmer Braren\inst{2} \and
Daniel Rueckert\inst{1,2,4} 
}
\authorrunning{A. Dima et al.}
\institute{
	School of Computation, Information and Technology, Technical University of Munich, Germany \and
	School of Medicine, Klinikum Rechts der Isar, Technical University of Munich, Germany \and
	Department of Quantitative Biomedicine, University of Zurich, Switzerland \and
	Department of Computing, Imperial College London, United Kingdom\\
	\email{alina.dima@tum.de}
}
\maketitle
%
\begin{abstract}
Automated segmentation of the blood vessels in 3D volumes is an essential step for the quantitative diagnosis and treatment of many vascular diseases.
3D vessel segmentation is being actively investigated in existing works, mostly in deep learning approaches.
However, training 3D deep networks requires large amounts of manual 3D annotations from experts, which are laborious to obtain. This is especially the case for 3D vessel segmentation, as vessels are sparse yet spread out over many slices and disconnected when visualized in 2D slices.
In this work, we propose a novel method to segment the 3D peripancreatic arteries \textbf{solely from one annotated 2D projection per training image} with depth supervision.
We perform extensive experiments on the segmentation of peripancreatic arteries on 3D contrast-enhanced CT images and demonstrate how well we capture the rich depth information from 2D projections.
We demonstrate that by annotating a single, randomly chosen projection for each training sample, we obtain comparable performance to annotating multiple 2D projections, thereby reducing the annotation effort.
Furthermore, by mapping the 2D labels to the 3D space using depth information and incorporating this into training, we almost close the performance gap between 3D supervision and 2D supervision.
Our code is available at: \href{https://github.com/alinafdima/3Dseg-mip-depth}{https://github.com/alinafdima/3Dseg-mip-depth}.


\end{abstract}

\section{Introduction}
Automated segmentation of blood vessels in 3D medical images is a crucial step for the diagnosis and treatment of many diseases, where the segmentation can aid in visualization, help with surgery planning, be used to compute biomarkers, and further downstream tasks.
Automatic vessel segmentation has been extensively studied, both using classical computer vision algorithms \cite{luboz2005segmentation} such as vesselness filters~\cite{frangi1998multiscale}, or more recently with deep learning~\cite{chen2021retinal,ciecholewski2021computational,isensee2021nnu,tetteh2020deepvesselnet,shi2019intracranial,dima2021segmentation}, where state-of-the-art performance has been achieved for various vessel structures.
Supervised deep learning typically requires large, well-curated training sets, which are often laborious to obtain. This is especially the case for 3D vessel segmentation. 

Manually delineating 3D vessels typically involves visualizing and annotating a 3D volume through a sequence of 2D cross-sectional slices, which is not a good medium for visualizing 3D vessels.
This is because often only the cross-section of a vessel is visible in a 2D slice.
In order to segment a vessel, the annotator has to track the cross-section of that vessel through several adjacent slices, which is especially tedious for curved or branching vessel trees.
Projecting 3D vessels to a 2D plane allows for the entire vessel tree to be visible within a single 2D image, providing a more robust representation and potentially alleviating the burden of manual annotation.
Kozinski \emph{et al.}~\cite{kozinski2020tracing} propose to annotate up to three maximum intensity projections (MIP) for the task of centerline segmentation~\cite{kozinski2020tracing}, obtaining results comparable to full 3D supervision.
Compared to centerline segmentation, where the vessel diameter is disregarded, training a 3D vessel segmentation model from 2D annotations poses additional segmentation-specific challenges, as 2D projections only capture the outline of the vessels, providing no information about their interior.
Furthermore, the axes of projection are crucial for the model's success, given the sparsity of information in 2D annotations.

To achieve 3D vessel segmentation with only 2D supervision from projections, we first investigate which viewpoints to annotate in order to maximize segmentation performance.
We show that it is feasible to segment the full extent of vessels in 3D images with high accuracy by annotating only a single randomly-selected 2D projection per training image.
This approach substantially reduces the annotation effort, even compared to works training only on 2D projections.
Secondly, by mapping the 2D annotations to the 3D space using the depth of the MIPs, we obtain a partially segmented 3D volume that can be used as an additional supervision signal.
We demonstrate the utility of our method on the challenging task of peripancreatic arterial segmentation on contrast-enhanced arterial-phase computed tomography (CT) images, which feature large variance in vessel diameter.
Our contribution to 3D vessel segmentation is three-fold:
\begin{itemize}
    \item [$\circ$] Our work shows that highly accurate automatic segmentation of 3D vessels can be learned by annotating single MIPs.
    \item [$\circ$] Based on extensive experimental results, we determine that the best annotation strategy is to label randomly selected viewpoints, while also substantially reducing the annotation cost.
    \item [$\circ$] By incorporating additional depth information obtained from 2D annotations at no extra cost to the annotator, we almost close the gap between 3D supervision and 2D supervision.
\end{itemize}

\section{Related Work}

\myparagraph{Learning from weak annotations.}
Weak annotations have been used in deep learning segmentation to reduce the annotation effort through cheaper, less accurate, or sparser labeling~\cite{tajbakhsh2020embracing}.
Bai \emph{et al.}~\cite{bai2018recurrent} learn to perform aortic image segmentation by sparsely annotating only a subset of the input slices.
Multiple instance learning approaches bin pixels together by only providing labels at the bin level.
Jia \emph{et al.}~\cite{jia2017constrained} use this approach to segment cancer on histopathology images successfully.
Annotating 2D projections for 3D data is another approach to using weak segmentation labels, which has garnered popularity recently in the medical domain.
Bayat \emph{et al.}~\cite{bayat2020inferring} propose to learn the spine posture from 2D radiographs, while Zhou \emph{et al.} \cite{zhou2019semi} use multi-planar MIPs for multi-organ segmentation of the abdomen. 
Kozinski \emph{et al.}\cite{kozinski2020tracing} propose to segment vessel centerlines using as few as 2-3 annotated MIPs.
Chen \emph{et al.} \cite{chen20233d} train a vessel segmentation model from unsupervised 2D labels transferred from a publicly available dataset, however, there is still a gap to be closed between unsupervised and supervised model performance.
Our work uses weak annotations in the form of annotations of 2D MIPs for the task of peripancreatic vessel segmentation, where we attempt to reduce the annotation cost to a minimum by only annotating a single projection per training input without sacrificing performance.

\myparagraph{Incorporating depth information.}
Depth is one of the properties of the 3D world.
Loss of depth information occurs whenever 3D data is projected onto a lower dimensional space.
In natural images, depth loss is inherent through image acquisition, therefore attempts to recover or model depth have been employed for 3D natural data.
For instance, Fu \emph{et al.}~\cite{fu2022panoptic} use neural implicit fields to semantically segment images by transferring labels from 3D primitives to 2D images.
Lawin \emph{et al.}~\cite{lawin2017deep} propose to segment 3D point clouds by projecting them onto 2D and training a 2D segmentation network.
At inference time, the predicted 2D segmentation labels are remapped back to the original 3D space using the depth information.
In the medical domain, depth information has been used in volume rendering techniques~\cite{drebin1988volume} to aid with visualization, but it has so far not been employed when working with 2D projections of 3D volumes to recover information loss.
We propose to do the conceptually opposite approach from Lawin \emph{et al.}~\cite{lawin2017deep}, by projecting 3D volumes onto 2D to facilitate and reduce annotation.
We use depth information to map the 2D annotations to the original 3D space at annotation time and generate partial 3D segmentation volumes, which we incorporate in training as an additional loss term.

\section{Methodology}

\newcommand{\R}{\mathbb{R}}
\newcommand{\qq}{\hspace{.25em}}
\newcommand{\proj}{\emph{mip}}
\newcommand{\depth}{\emph{D}}
\newcommand{\anno}{\emph{A}}
\newcommand{\ray}{\emph{r}}
\newcommand{\img}{\emph{I}}
\newcommand{\pfw}{\emph{p}^{fw}}
\newcommand{\pbw}{\emph{p}^{bw}}
\newcommand{\zfw}{\emph{z}^{fw}}
\newcommand{\zbw}{\emph{z}^{bw}}

\begin{figure}[t!]
	\centering
	\includegraphics[width=\linewidth, height=.45\linewidth]{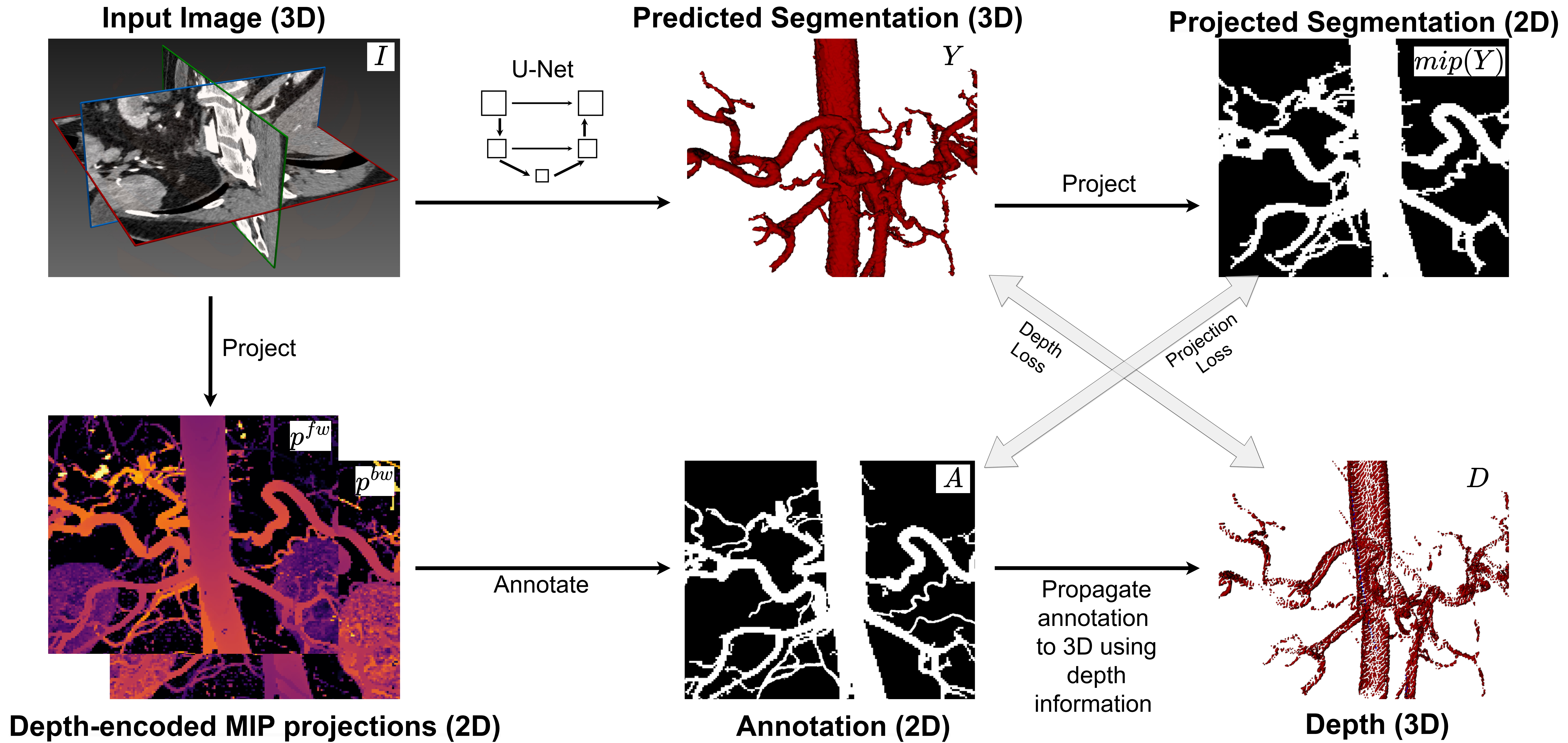}
	\caption{Method overview. 
    We train a 3D network to segment vessels from 2D annotations.
    Given an input image $\img$, depth-encoded MIPs $\pfw, \pbw$ are generated by projecting the input image to 2D.
    2D binary labels $\anno$ are generated by annotating one 2D projection per image.
    The 2D annotation is mapped to the 3D space using the depth information, resulting in a partially labeled 3D volume $\depth$. 
    During training, both 2D annotations and 3D depth maps are used as supervision signals in a combined loss, which uses both predicted 3D segmentation $Y$ and its 2D projection $mip(Y)$.
 }
	\label{fig:main}
\end{figure}

\myparagraph{Overview.}
The maximum intensity projection (MIP) of a 3D volume \\ \mbox{$\img \in \R^{N_x\times N_y \times N_z}$} is defined as the highest intensity along a given axis:
\begin{equation}
\mbox{$\proj(x, y) = \max_{z} \qq \img (x, y, z) \in \R^{N_x\times N_y}$}.
\end{equation}
For simplicity, we only describe MIPs along the z-axis, but they can be performed on any image axis.

Exploiting the fact that arteries are hyperintense in arterial phase CTs, we propose to annotate MIPs of the input volume for binary segmentation.
The hyperintensities of the arteries ensures their visibility in the MIP, while additional processing removes most occluding nearby tissue (Section \ref{section:dataset}).

Given a binary 2D annotation of a MIP $A\in \{0,1\}^{N_x \times N_y}$, we map the foreground pixels in A to the original 3D image space.
This is achieved by using the first and last $z$ coordinates where the maximum intensity is observed along any projection ray.
Owing to the fact that the vessels in the abdominal cavity are relatively sparse in 2D projections and most of the occluding tissue is removed in postprocessing, this step results in a fairly complete surface of the vessel tree.
Furthermore, we can partially fill this surface volume, resulting in a 3D depth map $\depth$, which is a partial segmentation of the vessel tree.
We use the 2D annotations as well as the depth map to train a 3D segmentation network in a weakly supervised manner.

An overview of our method is presented in Figure \ref{fig:main}.
In the following, we describe these components and how they are combined to train a 3D segmentation network in more detail.

\newcommand{\imW}{.24\linewidth}
\newcommand{\imH}{.16\linewidth}
\begin{figure}[t]
	\centering
	\begin{tabular}{cccc}
		\includegraphics[width=\imW, height=\imH]{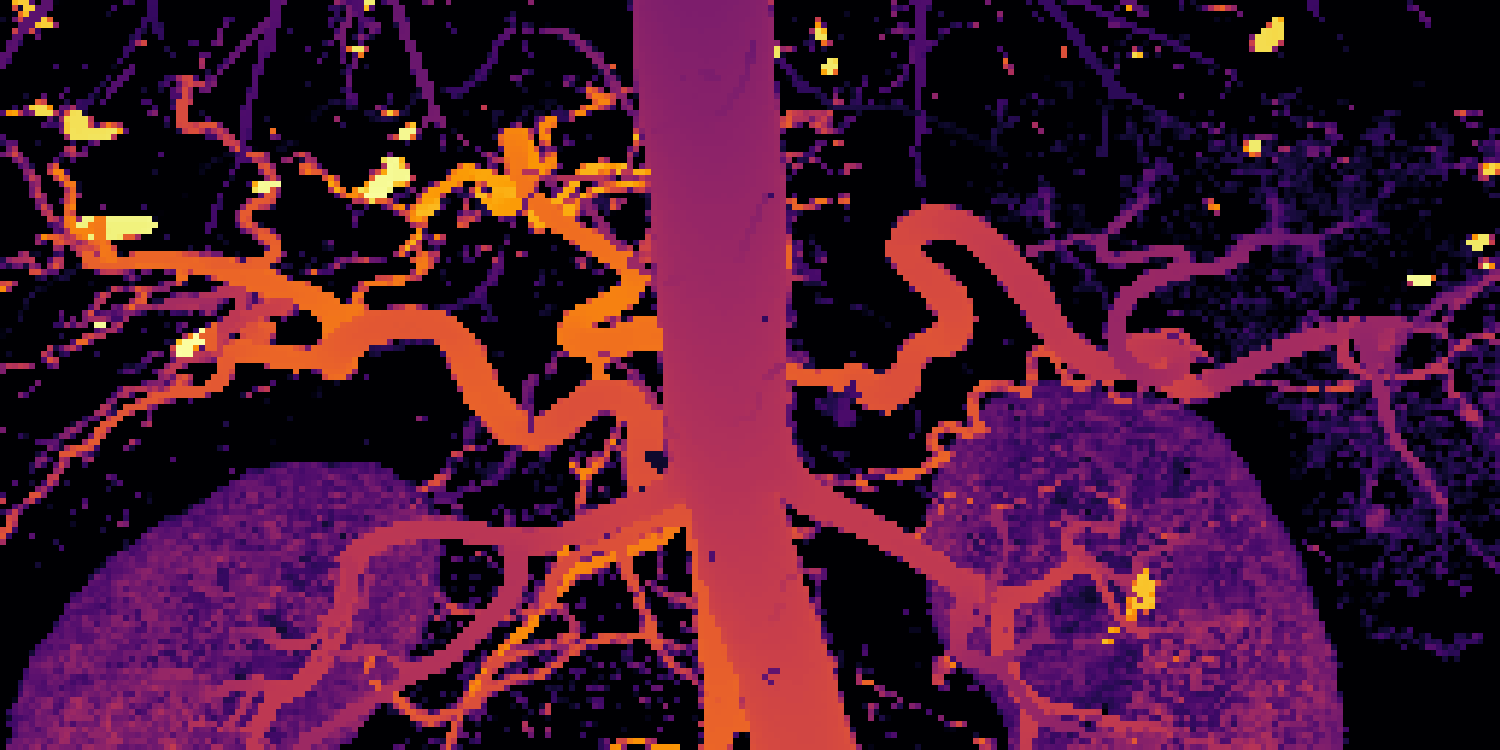} & 
		\includegraphics[width=\imW, height=\imH]{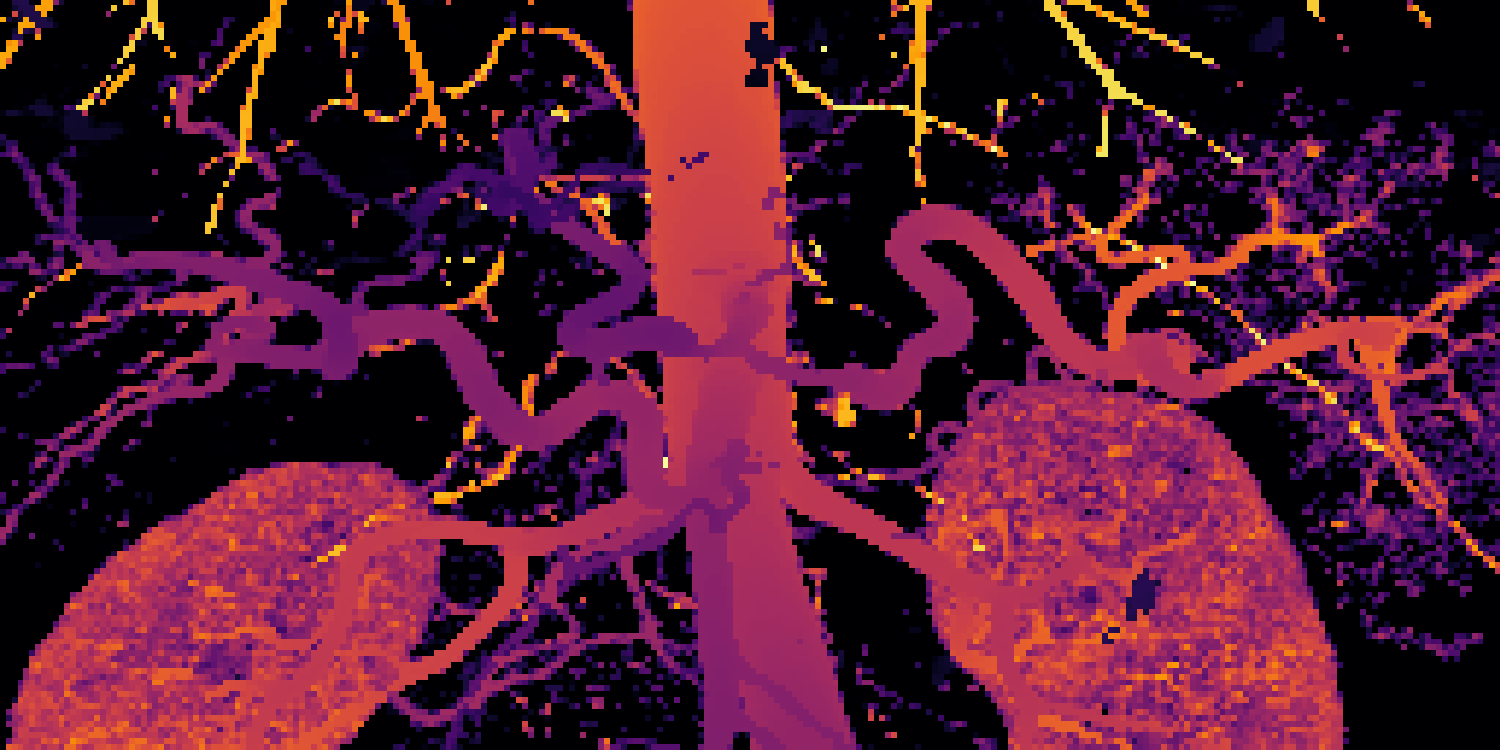} &
		\includegraphics[width=\imW, height=\imH]{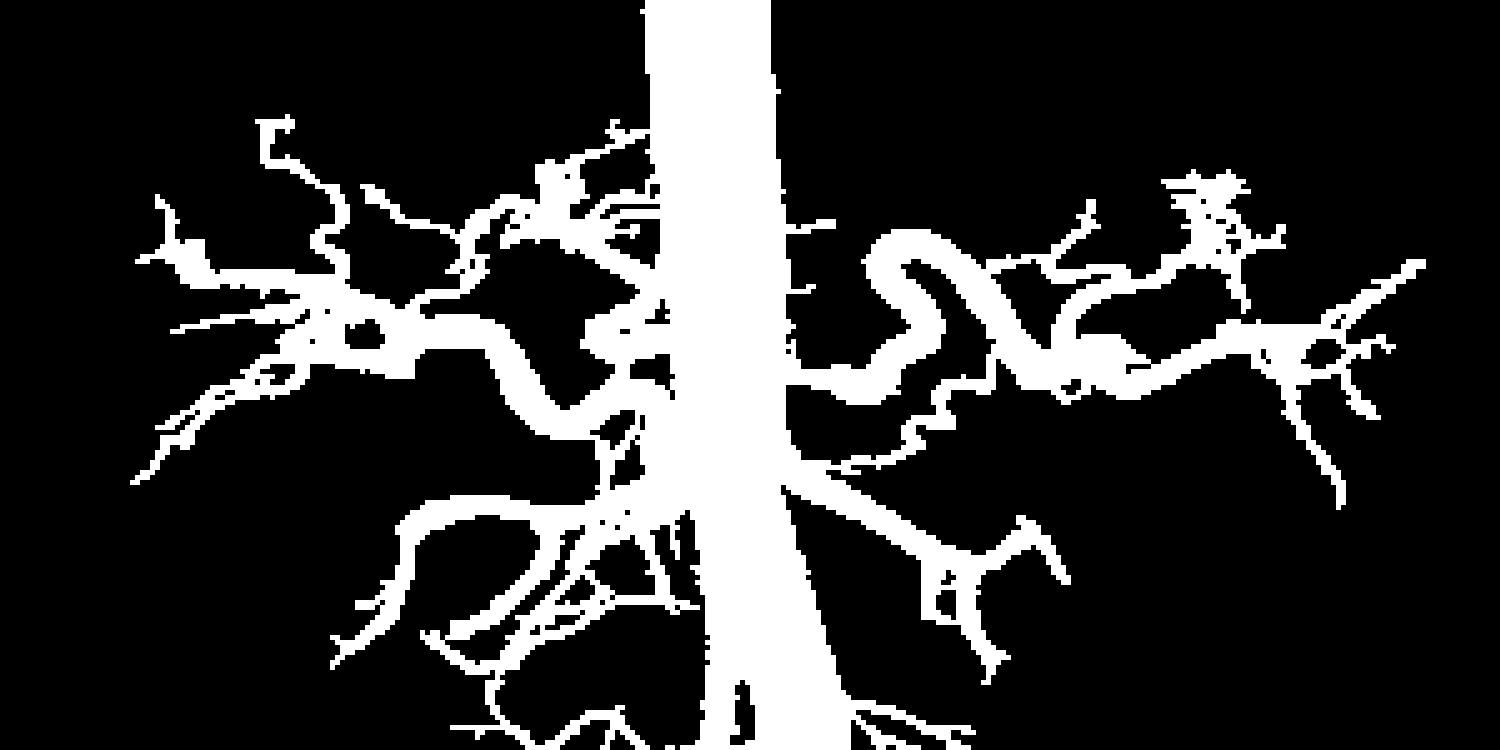} & 
        \includegraphics[width=\imW, height=\imH]{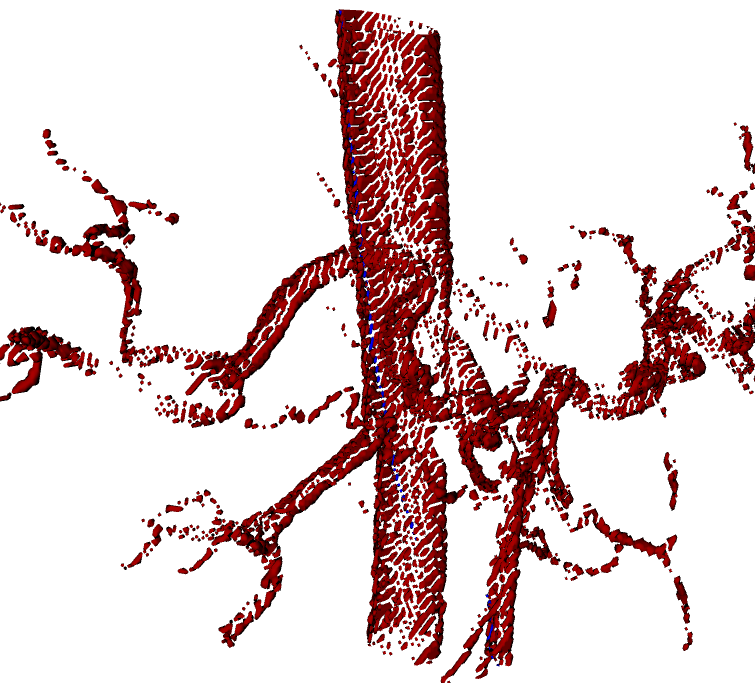} \\
  		\scriptsize{(a)} & \scriptsize{(b)} & \scriptsize{(c)} & \scriptsize{(d)}  \\
		\includegraphics[width=\imW, height=\imH]{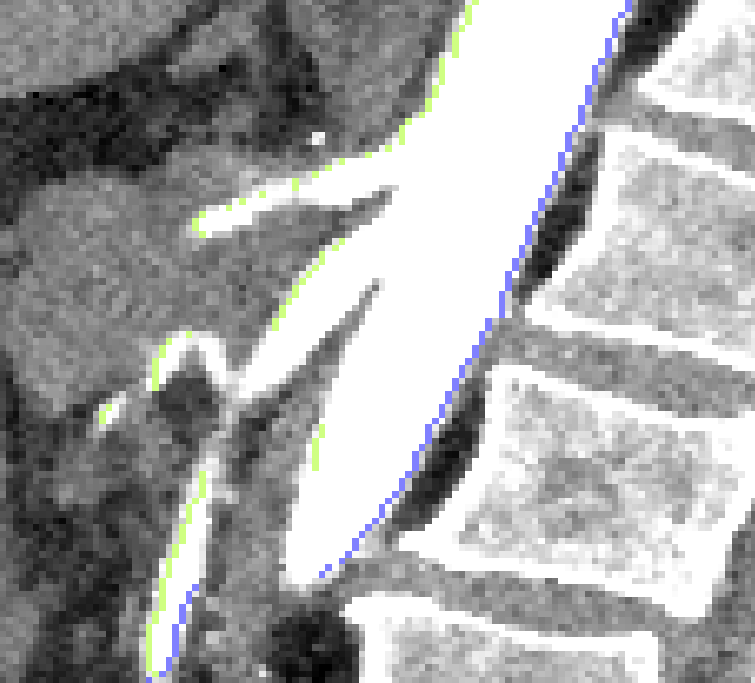} & 
		\includegraphics[width=\imW, height=\imH]{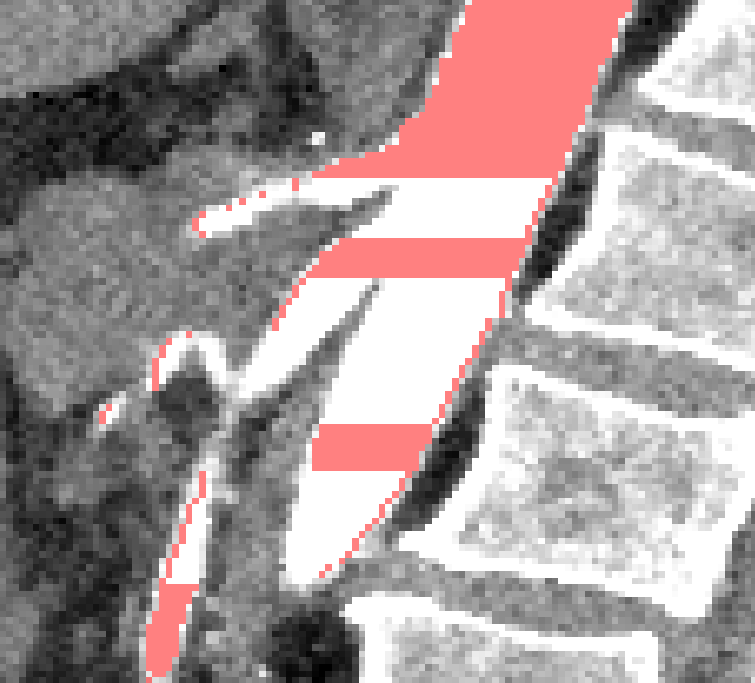} & 
		\includegraphics[width=\imW, height=\imH]{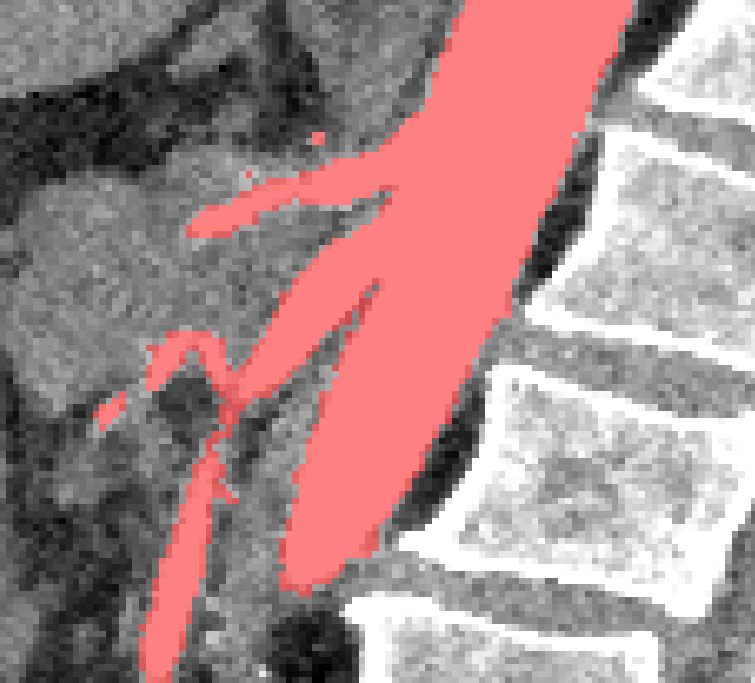} & 
		\includegraphics[width=\imW, height=\imH]{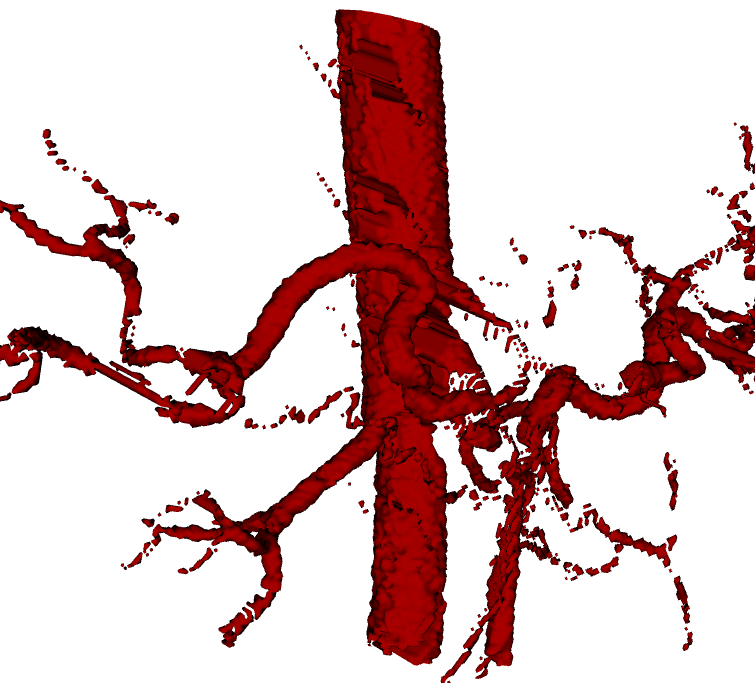} \\
  		\scriptsize{(e)} & \scriptsize{(f)} & \scriptsize{(g)} & \scriptsize{(h)} 

	\end{tabular}
	\caption{
    Example depth-enhanced MIP using 
    (a) forward depth $\zfw$ and 
    (b) backward depth $\zbw$ visualized in color;
    (c) binary 2D annotation;
    a slice view from a 3D volume illustrating: 
    (e) the forward -- in {\color{green}green} --  and backward depth -- in {\color{blue}blue} -- , 
    (f) the depth map, 
    (g) 3D ground truth;
    volume rendering of 
    (h) the depth map and 
    (d) the depth map with only forward and backward depth pixels.
    The input images are contrast-enhanced.
 }
	\label{fig:mip}
\end{figure}

\myparagraph{Depth information.}
We can view MIP as capturing the intensity of the brightest pixel along each ray $\ray_{xy} \in\R^{N_z}$, where $\ray_{xy}(z) = \img(x, y, z)$. 
Along each projection ray, we denote the first and last $z$ coordinates which have the same intensity as the MIP to be the forward depth $\zfw = \argmax_{z} \qq \img (x, y, z)$ and backward depth $\zbw = \argmin_{z} \qq \img (x, y, z)$.
This information can be utilized for the following: 
(1) enhancing the MIP visualization, or (2) providing a way to map pixels from the 2D MIP back to the 3D space (depth map).
The reason why the maximum intensity is achieved multiple times along a ray is because our images are clipped, which removes a lot of the intensity fluctuations.

\myparagraph{Depth-enhanced MIP.}
We encode depth information into the MIPs by combining the MIP with the forward and backward depth respectively, in order to achieve better depth perception during annotation:
$\pfw = \sqrt{\proj} \cdot \zfw$ defines the forward projection, while
$\pbw = \sqrt{\proj} \cdot \zbw$ defines the backward projection.
Figure \ref{fig:mip} showcases (a) forward and (b) backward depth encoded MIPs.


\myparagraph{Depth map generation.}
Foreground pixels from the 2D annotations are mapped to the 3D space by combining a 2D annotation with the forward and backward depth, resulting in a 3D partial vessel segmentation:
\begin{topbot}
\begin{enumerate}
    \item[\textbf{1.}] Create an empty 3D volume $\depth \in \R^{N_x\times N_y \times N_z}$.
    \item[\textbf{2.}] For each foreground pixel in the annotation $\anno$ at location $(x, y)$, we label $(x, y, \zfw)$ and $(x, y, \zbw)$ as foreground pixels in $\depth$.
    \item[\textbf{3.}] If the fluctuation in intensity between $\zfw$ and $\zbw$ along the ray $\ray_{xy}$ is below a certain threshold in the source image \img, the intermediate pixels are also labeled as foreground in $\depth$.
\end{enumerate}
\end{topbot}

\myparagraph{Training loss.}
We train a 3D segmentation network to predict 3D binary vessel segmentation given a 3D input volume using 2D annotations.
Our training set $\mathcal{D}_{tr}(\img, \anno, \depth)$ consists of 3D volumes $\img$ paired with 2D annotations $\anno$ and their corresponding 3D depth maps $\depth$.
Given the 3D network output $\emph{Y}=\theta(\img)$, we minimize the following loss during training:
\begin{equation}
	\mathcal{L}(\emph{Y}) = 
		\alpha \cdot \mathcal{CE}(\anno, \qq mip(\emph{Y})) + 
		(1 - \alpha)\cdot \mathcal{CE}(\depth, \qq \emph{Y}) \cdot \depth,
\end{equation}
where $\alpha \in [0, 1]$.
Our final loss is a convex combination between: 
\textbf{(a)} the cross-entropy($\mathcal{CE}$) of the network output projected to 2D and the 2D annotation, as well as  
\textbf{(b)} the cross-entropy between the network output and the depth map, but only applied to positive pixels in the depth map.
Notably, the 2D loss constrains the shape of the vessels, while the depth loss promotes the segmentation of the vessel interior.

\section{Experimental Design}
\label{experiments}

\myparagraph{Dataset.} 
\label{section:dataset}
We use an in-house dataset of contrast-enhanced abdominal computed tomography images (CTs) in the arterial phase to segment the peripancreatic arteries~\cite{dima2021segmentation}.
The cohort consists of 141 patients with pancreatic ductal adenocarcinoma, of an equal ratio of male to female patients.
Given a 3D arterial CT of the abdominal area, we automatically extract the vertebrae~\cite{loffler2020vertebral,sekuboyina2021verse} and semi-automatically extract the ribs, which have similar intensities as arteries in arterial CTs and would otherwise occlude the vessels.
In order to remove as much of the cluttering surrounding tissue and increase the visibility of the vessels in the projections, the input is windowed so that the vessels appear hyperintense.
Details of the exact preprocessing steps can be found in Table 2 of the supplementary material.
The dataset contains binary 3D annotations of the peripancreatic arteries carried out by two radiologists, each having annotated half of the dataset.
The 2D annotations we use in our experiments are projections of these 3D annotations.
For more information about the dataset, see~\cite{dima2021segmentation}.

\myparagraph{Image augmentation and transformation.}
As the annotations lie on a 2D plane, 3D spatial augmentation cannot be used due to the information sparsity in the ground truth.
Instead, we apply an invertible transformation $\mathcal{T}$ to the input volume and apply the inverse transformation $\mathcal{T}^{-1}$ to the network output before applying the loss, such that the ground truth need not be altered. A detailed description of the augmentations and transformations used can be found in Table 1 in the supplementary material.

\myparagraph{Training and evaluation.} 
We use a 3D U-Net \cite{ronneberger2015u} with four layers as our backbone, together with Xavier initialization \cite{he2015delving}.
A diagram of the network architecture can be found in Figure 2 in the supplementary material.
The loss weight $\alpha$ is tuned at $0.5$, as this empirically yields the best performance.
Our experiments are averaged over 5-fold cross-validation with 80 train samples, 20 validation samples, and a fixed test set of 41 samples.
The network initialization is different for each fold but kept consistent across different experiments run on the same fold.
This way, both data variance and initialization variance are accounted for through cross-validation.
To measure the performance of our models, we use the Dice score, precision, recall, and mean surface distance (MSD).
We also compute the skeleton recall as the percentage of the ground truth skeleton pixels which are present in the prediction.

\setlength{\tabcolsep}{3pt}
\begin{table}[t]
\centering
\caption{%
Viewpoint ablation. 
We compare models trained on single random viewpoints (VPs) with ($+$D) or without ($-$D) depth against fixed viewpoint baselines without depth and full 3D supervision. 
We distinguish between model selection based on 2D annotations vs. 3D annotations on the validation set.
The best-performing models for each model selection (2D \emph{vs.} 3D) are highlighted in bold.
}
\scalebox{0.82}{
\begin{tabular}{c c c c c c c}
\toprule
    \multirow{2}*{Experiment} & Model & \multirow{2}*{Dice $\uparrow$} & \multirow{2}*{Precision $\uparrow$} & \multirow{2}*{Recall $\uparrow$} & Skeleton & \multirow{2}*{MSD  $\downarrow$}\\
     & Selection & & & & Recall $\uparrow$ & \\
	\toprule
	3D & 3D & $\mathbf{92.18 \pm 0.35}$ & $\mathbf{93.86 \pm 0.81}$ & $90.64 \pm 0.64$ & $76.04 \pm 4.51$ & $1.15 \pm 0.11$ \\
	\midrule
	fixed 3VP & 3D & $92.02 \pm 0.52$ & $93.05 \pm 0.61$ & $91.13 \pm 0.79$ & $\mathbf{78.61 \pm 1.52}$ & $1.13 \pm 0.11$ \\
	fixed 2VP & 3D & $91.29 \pm 0.78$ & $91.46 \pm 2.13$ & $\mathbf{91.37 \pm 1.45}$ & $78.51 \pm 2.78$ & $\mathbf{1.13 \pm 0.09}$ \\
	\midrule
	fixed 3VP & 2D & $90.78 \pm 1.30$ & $90.66 \pm 1.30$ & $91.18 \pm 3.08$ & $81.77 \pm 2.13$ & $1.16 \pm 0.13$ \\
	fixed 2VP & 2D & $90.22 \pm 1.19$ & $88.16 \pm 2.86$ & $92.74 \pm 1.63$ & $\mathbf{82.18 \pm 2.47}$ & $1.14 \pm 0.09$ \\
	fixed 1VP & 2D & $60.76 \pm 24.14$ & $50.47 \pm 23.21$ & $92.52 \pm 3.09$ & $81.19 \pm 2.39$ & $2.96 \pm 3.15$ \\
	\midrule
	random 1VP$-$D & 2D & $91.29 \pm 0.81$ & $\mathbf{91.42 \pm 0.92}$ & $91.45 \pm 1.00$ & $80.16 \pm 2.35$ & $\mathbf{1.13 \pm 0.04}$ \\
	random 1VP$+$D & 2D & $\mathbf{91.69 \pm 0.48}$ & $90.77 \pm 1.76$ & $\mathbf{92.79 \pm 0.95}$ & $81.27 \pm 2.02$ & $1.15 \pm 0.11$ \\
	\bottomrule
\end{tabular}
}
\label{table:main}
\end{table}

\section{Results}

\noindent\textbf{The effectiveness of 2D projections and depth supervision.} We compare training using single random viewpoints with and without depth information against baselines that use more supervision.
Models trained on full 3D ground truth represent the upper bound baseline, which is very expensive to annotate.
We implement~\cite{kozinski2020tracing} as a baseline on our dataset, training on up to 3 fixed orthogonal projections.
We distinguish between models selected according to the 2D performance on the validation set (2D) which is a fair baseline, and models selected according to the 3D performance on the validation set (3D), which is an unfair baseline as it requires 3D annotations on the validation set.
With the exception of the single fixed viewpoint baselines where the models have the tendency to diverge towards over- or segmentation, we perform binary hole-filling on the output of all of our other models, as producing hollow objects is a common under-segmentation issue.

In Table \ref{table:main} we compare our method against the 3D baseline, as well as baselines trained on multiple viewpoints. 
We see that by using \textbf{depth information} paired with training using a single random viewpoint per sample performs almost at the level of models trained on 3D labels, at a very small fraction of the annotation cost.
The depth information also reduces model variance compared to the same setup without depth information.
Even without depth information, training the model on single \textbf{randomly} chosen viewpoints offers a robust training signal that the Dice score is on par with training on 2 fixed viewpoints under ideal model selection at only half the annotation cost.
Randomly selecting viewpoints for training acts as powerful data augmentation, which is why we are able to obtain performance comparable to using more fixed viewpoints.
Under ideal 3D-based model selection, three views would come even closer to full 3D performance; however, with realistic 2D-based model selection, fixed viewpoints are more prone to diverge.
This occurs because sometimes 2D-based model selection favors divergent models which only segment hollow objects, which cannot be fixed in postprocessing.
Single fixed viewpoints contain so little information on their own that models trained on such input fail to learn how to segment the vessels and generally converge to over-segmenting in the blind spots in the projections.
We conclude that using random viewpoints is not only helpful in reducing annotation cost but also decreases model variance.

In terms of other metrics, randomly chosen projection viewpoints with and without depth improve both recall and skeleton recall even compared to fully 3D annotations, while generally reducing precision.
We theorize that this is because the dataset itself contains noisy annotations and fully supervised models better overfit to the type of data annotation, whereas our models converge to following the contrast and segmenting more vessels, which are sometimes wrongfully labeled as background in the ground truth.
MSD are not very telling in our dataset due to the noisy annotations and the nature of vessels, as an under- or over-segmented vessel branch can quickly translate into a large surface distance.

\myparagraph{The effect of dataset size.}
We vary the size of the training set from $|\mathcal{D}_{tr}| = 80$ to as little as $|\mathcal{D}_{tr}| = 10$ samples, while keeping the size of the validation and test sets constant, and train models on single random viewpoints.

In Table \ref{table:dsize}, we compare single random projections trained with and without depth information at varying dataset sizes to ilustrate the usefulness of the depth information with different amounts of training data.
Our depth loss offers consistent improvement across multiple dataset sizes and reduces the overall performance variance.
The performance boost is noticeable across the board, the only exception being precision.
The smaller the dataset size is, the greater the performance boost from the depth.
We perform a Wilcoxon rank-sum statistical test comparing the individual sample predictions of the models trained at various dataset sizes with single random orthogonal viewpoints with or without depth information, obtaining a statistically significant (p-value of $<0.0001$). We conclude that the depth information complements the segmentation effectively.

\setlength{\tabcolsep}{3pt}
\begin{table}[t]
\small
\centering
\caption{Dataset size ablation. We vary the training dataset size $|\mathcal{D}_{tr}|$ and compare models trained on single random viewpoints, with or without depth. Best performing models in each setting are highlighted.}
\scalebox{0.85}{
\begin{tabular}{c c c c c c c}
\toprule
    \multirow{2}*{$|\mathcal{D}_{tr}|$} & \multirow{2}*{Depth} & 
        \multirow{2}*{Dice $\uparrow$} & \multirow{2}*{Precision $\uparrow$} & \multirow{2}*{Recall $\uparrow$} & 
        Skeleton & \multirow{2}*{MSD  $\downarrow$} \\
     & & & & & Recall $\uparrow$ & \\
	\toprule
	10 & $-$D & $86.03 \pm 2.94$ & $88.23 \pm 2.58$ & $84.81 \pm 6.42$ & $78.25 \pm 2.20$ & $1.92 \pm 0.55$ \\
	10 & $+$D & $\mathbf{89.06 \pm 1.20}$ & $\mathbf{88.55 \pm 1.73}$ & $\mathbf{89.91 \pm 1.29}$ & $\mathbf{78.95 \pm 3.62}$ & $\mathbf{1.80 \pm 0.28}$ \\
	\midrule
	20 & $-$D & $88.22 \pm 3.89$ & $\mathbf{90.26 \pm 1.64}$ & $86.74 \pm 6.56$ & $80.78 \pm 1.66$ & $1.44 \pm 0.20$ \\
	20 & $+$D & $\mathbf{90.51 \pm 0.38}$ & $89.84 \pm 0.90$ & $\mathbf{91.50 \pm 1.23}$ & $\mathbf{80.00 \pm 1.95}$ & $\mathbf{1.33 \pm 0.16}$ \\
	\midrule
	40 & $-$D & $88.07 \pm 2.34$ & $\mathbf{89.09 \pm 2.01}$ & $87.62 \pm 4.43$ & $78.38 \pm 2.39$ & $1.38 \pm 0.10$ \\
	40 & $+$D & $\mathbf{90.21 \pm 0.89}$ & $89.08 \pm 2.89$ & $\mathbf{91.82 \pm 2.11}$ & $\mathbf{79.16 \pm 2.36}$ & $\mathbf{1.24 \pm 0.14}$ \\
	\midrule
	80 & $-$D & $91.29 \pm 0.81$ & $\mathbf{91.42 \pm 0.92}$ & $91.45 \pm 1.00$ & $80.16 \pm 2.35$ & $\mathbf{1.13 \pm 0.04}$ \\
	80 & $+$D & $\mathbf{91.69 \pm 0.48}$ & $90.77 \pm 1.76$ & $\mathbf{92.79 \pm 0.95}$ & $\mathbf{81.27 \pm 2.02}$ & $1.15 \pm 0.11$ \\
	\bottomrule
    \end{tabular}
    }
    \label{table:dsize}
\end{table}

%
\section{Conclusion}
In this work, we present an approach for 3D segmentation of peripancreatic arteries using very sparse 2D annotations.
Using a labeled dataset consisting of single, randomly selected, orthogonal 2D annotations for each training sample and additional depth information obtained at no extra cost, we obtain accuracy almost on par with fully supervised models trained on 3D data at a mere fraction of the annotation cost.
Limitations of our work are that the depth information relies on the assumption that the vessels exhibit minimal intensity fluctuations within local neighborhoods, which might not hold on other datasets, where more sophisticated ray-tracing methods would be more effective in locating the front and back of projected objects.
Furthermore, careful preprocessing is performed to eliminate occluders, which would limit its transferability to datasets with many occluding objects of similar intensities.
Further investigation is needed to quantify how manual 2D annotations compare to our 3D-derived annotations, where we expect occluders to affect the annotation process.


%
\bibliographystyle{splncs04}
\bibliography{bibliography}

\end{document}